\documentclass{article}
\usepackage{spconf}
\usepackage{graphicx}
\usepackage{float}
\usepackage{amsmath}
\usepackage{colortbl,hhline}
\usepackage{booktabs}
\usepackage{diagbox}
\usepackage{comment}
\usepackage[font=small,skip=4pt]{caption}
\usepackage{bbold}
\usepackage{graphicx}
\usepackage{url}
\usepackage{comment}

\graphicspath{{images/}}

\title{The Fibonacci  Network:
A Simple Alternative for Positional Encoding}
\name{Yair Bleiberg \& Michael Werman}

\address{Department of Computer Science, The Hebrew University of Jerusalem}

\begin{document}

\maketitle
\begin{abstract}
     Coordinate-based Multi-Layer Perceptrons (MLPs) are known to have difficulty reconstructing high frequencies of the training data. A common solution to this problem is Positional Encoding (PE), which has become quite popular. However, PE has drawbacks. It has high-frequency artifacts and adds another hyper-hyperparameter, just like batch normalization and dropout do. We believe that under certain circumstances PE is not necessary, and a smarter construction of the network architecture together with a smart training method is sufficient to achieve similar results. In this paper, we show that very simple MLPs can quite easily output a frequency when given input of the half-frequency and quarter-frequency. Using this, we design a network architecture in blocks, where the input to each block is the output of the two previous blocks along with the original input. We call this a {\it Fibonacci Network}. By training each block on the corresponding frequencies of the signal, we show that Fibonacci Networks can reconstruct arbitrarily high frequencies.
\end{abstract}

\section{Introduction}

It has become  popular in the field of Deep Learning to use neural networks to reconstruct continuous approximations, replacing the standard approach of using discrete pixels, voxels, and meshes. Neural networks which receive coordinates as input, known as Coordinate-based MLPs, are now used in such diverse domains as images \cite{nguyen2015deep}, signed distance \cite{park2019deepsdf}, and audio \cite{szatkowski2022hypersound}. Coordinate-based MLPs offer a practical approach to reconstruct and perform tasks within such environments, including novel view synthesis \cite{mildenhall2020nerf,saito2019pifu,niemeyer2020differentiable}, shape representation \cite{chen2019learning,deng2022nasa}, and texture synthesis \cite{henzler2020learning}. These networks possess a significant advantage in terms of memory efficiency, as they can compress gigabytes of environmental data into a few megabytes of weight parameters. However, an identified limitation of coordinate-based neural networks with ReLU activations is their struggle to reconstruct high frequencies in low dimensions. This is known as spectral bias \cite{rahaman2019spectral}. In addressing this issue, a common solution employed is Positional Encoding. In this thesis, we aim to explore alternative network architectures for coordinate-MLPs that can reduce or eliminate the reliance on Positional Encoding. We introduce the Fibonacci Network, a block-based architecture, along with a unique training methodology. The motivation behind this unconventional architecture and the results achieved are detailed further in the subsequent sections.

\section{Related Work}

Positional Encoding can be traced back to Fourier feature mappings initially proposed by Rahimi and Recht \cite{rahimi2007random}. It gained popularity in the field of Natural Language Processing (NLP) with its application in Transformers \cite{vaswani2017attention}. Since the attention mechanism introduced there does not provide information about the positioning of a word in a sequence, a sinusoidal encoding of the position is  used to tell the Transformer where in the sequence the word is. The use of Positional Encoding has since been adapted for coordinate-MLPs. Tancik \cite{tancik2020fourier} utilized Neural Tangent Kernel (NTK) theory \cite{jacot2020neural,rahaman2019spectral}  to provide theoretical justification for positional encoding in coordinate-MLPs.  These results have been used to great effect in NeRF \cite{mildenhall2020nerf}. Zheng \cite{zheng2021rethinking} questioned the necessity of sinusoidal positional encoding, and use a framework of shifted basis functions to show that any encoding which both preserves the stable rank of the embedding matrix and the distance between embedded coordinates can be a candidate for positional encoding. 

\section{Background}

Positional Encoding trains the network on a function of the input, rather than only on the input itself. In the context of coordinate-MLPs, a typical embedding method involves using sine and cosine embeddings. For a network N and input $x\in\mathbb{R}^d$, training is performed on inputs of the form:

\[\sin(2^jx_i),\cos(2^jx_i)~|~ i\in[d]\] 
where d represents the dimension of the input. Another encoding technique, referred to as "Gaussian" by Tancik, yields slightly improved results:

\[(\sin(a^Tx),\cos(a^Tx)~| ~a_i \in N(0,\sigma))\]
where $\sigma$ is a hyperparameter. However, both encoding methods have a significant drawback – the choice of frequency spectrum employed in the encoding significantly affects its effectiveness. If the frequencies are too low, the encoding becomes ineffective, while high frequencies can introduce artifacts into the output. Since each target function has a unique frequency distribution, a hyperparameter sweep is necessary to determine the optimal encoding. Additionally, the second option mentioned above utilizes random frequencies, leading to potential performance variations between runs. In this study, we present an alternative  method that progressively reconstructs higher frequencies, capitalizing on interesting experimental findings, which are elaborated upon in subsequent sections.

\section{Leapfrogging Frequencies}
It is very difficult to train a fully-connected network on high frequencies. A question arises: if a network can reconstruct very nontrivial functions such as is done by NeRF with only the help of the input of high-frequency functions, should it be able to reconstruct high frequencies with the help of lower ones?
\\The answer is, surprisingly, Yes! For example, let's attempt to reconstruct the function $f(x)=\sin(512x)$.
As expected, a standard 1-D coordinate-MLP $N(\{x\})$ severely under fits.

\begin{figure*}[h]
    \centering
    \includegraphics[width=1.1\textwidth]{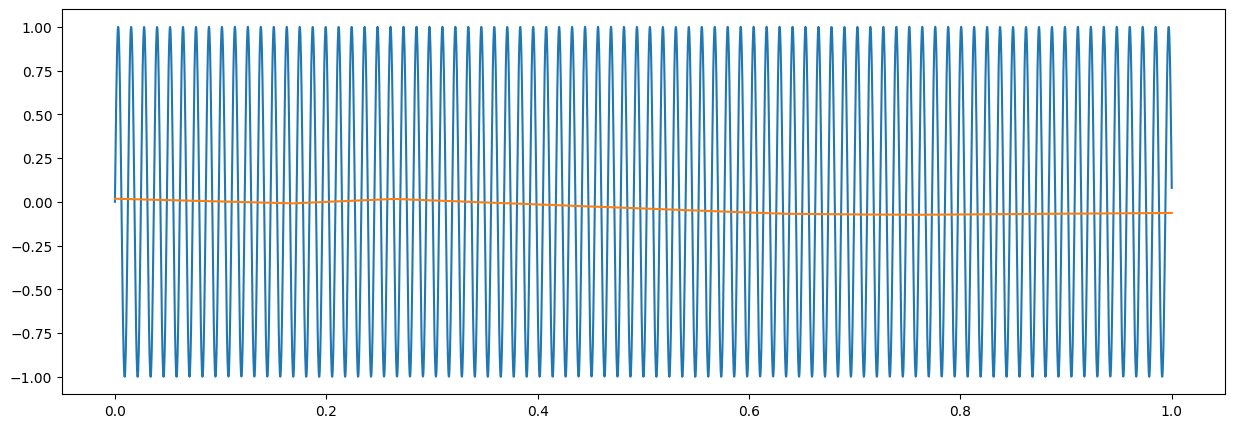}
    \caption{The MLP's reconstruction of sin(512x) (orange), compared with the ground truth (blue).}
    \label{fig:no_freq}
\end{figure*}

Now, when we provide the network with the half-frequency, $N(\{x,\sin(256x)\})$, the result is a severe overfit.

\begin{figure*}[h]
    \centering
    \includegraphics[width=1\textwidth]{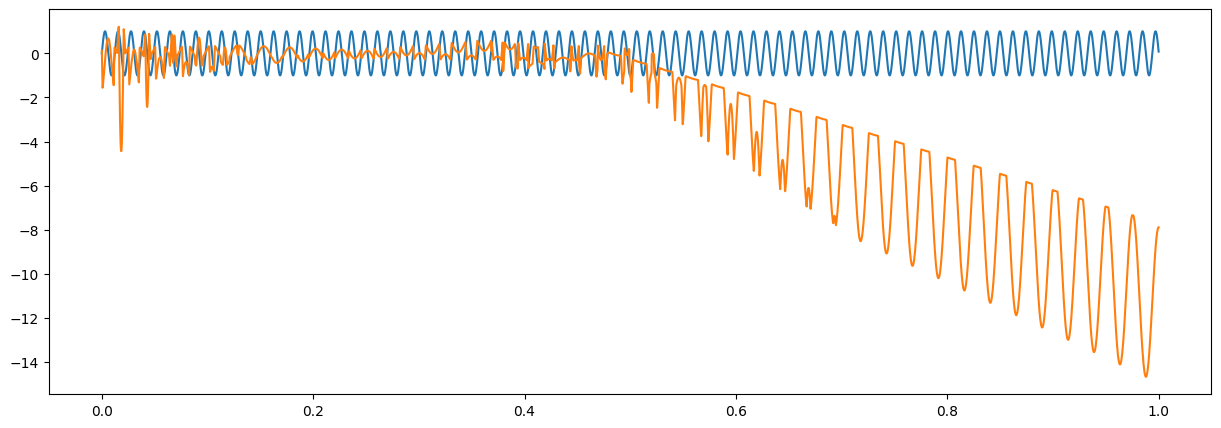}
    \caption{The MLP's reconstruction with both x and sin(256x) as input.}
    \label{fig:half_freq}
\end{figure*}

The real magic happens when the network is provided with both the half-frequency and the quarter-frequency, 
$N(\{x,\sin(128x),\sin(256x)\})$ . In this case, the reconstruction is nearly perfect, and the generalization as well.

\begin{figure*}[h]
    \includegraphics[width=0.91\textwidth]{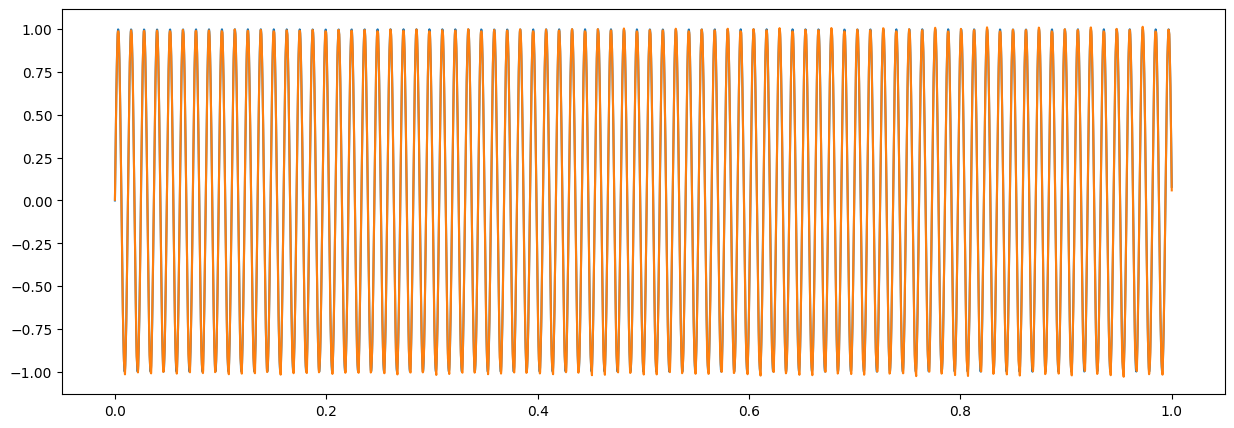}
    \caption{The MLP's reconstruction with x, sin(256x), and sin(128x) as input.
    }
    \label{fig:half_and_quarter_freq}
\end{figure*}

The above results are obtained with an extremely narrow and shallow network, with $width=8$ and $depth=2$.
The training data is likewise sparse, using 100 points uniformly sampled on the interval $[0,0.5]$. Notably, the last network is able to generalize nearly flawlessly to the interval $[0.5,1]$.

As we can see, when using the two previous frequencies, we not only obtain a much better fit but a much better generalization as well. We have not trained at all on the interval $[0.5,1]$, yet the two-frequency input is able to generalize nearly perfectly to the interval, whereas no frequency input and half-frequency input both fail severely.

What's going on? It appears that while the network struggles to reconstruct the high frequency with only the coordinate as input, it's able to find a semi-linear combination of the half and the quarter frequency is able to reach the desired output. Taking the trained half and quarter-frequencies network, if we look at  the output of the last layer of hidden neurons as a function of the input to the network, the neuron parametrization is some kind of periodic feature!

\begin{figure}[h]
    \includegraphics[width=0.45\textwidth]{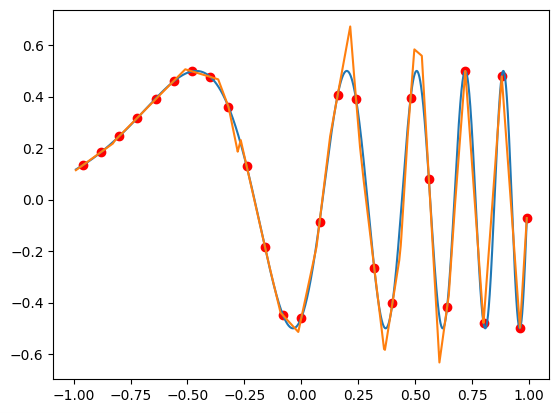}
    \caption{Taking the network from \cite{hertz2021sape} and training for more epochs. The training data are the red dots, the desired reconstructed signal is in blue, and our reconstruction is in orange.}
    \label{fig:sape_no_enc}
\end{figure}

\begin{figure*}[h]
    \includegraphics[width=0.82\textwidth]{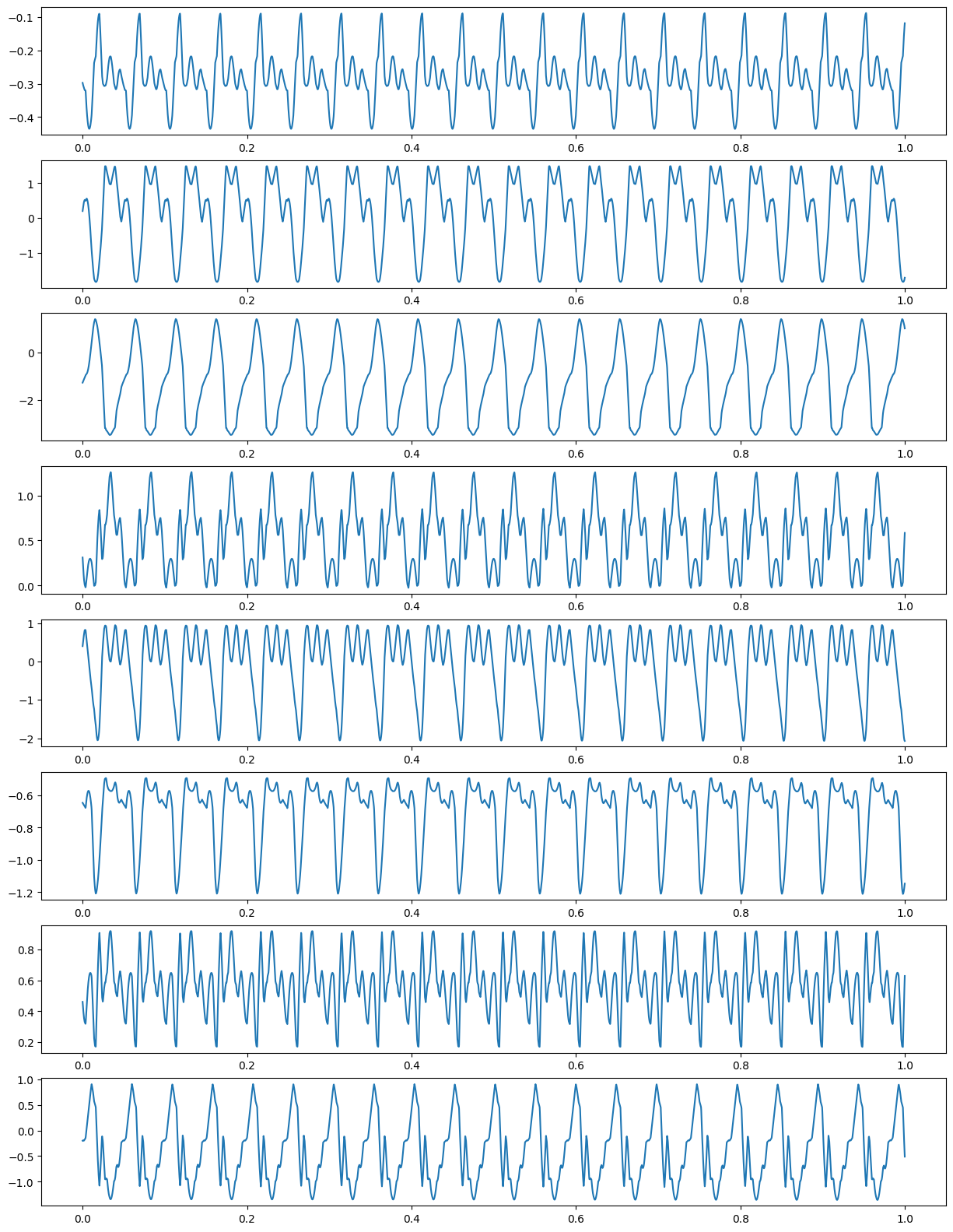}
    \caption{Output of hidden neurons as function of input to network $(x,\sin(256x),\sin(128x))$. Each plot is a separate neuron output.}
    \label{fig:neuron_parametrization}
\end{figure*}

Note that the output neurons are periodic, and generalize to the entire $[0,1]$ interval, not just the $[0,0.5]$ interval they were trained on.

\section{Fibonacci Network}
Leveraging this interesting result, we create a new network architecture: A series of simple neural blocks, where the input to each block is the output of the previous two blocks. The idea is that given a function with high frequencies that we wish to reconstruct, the earlier blocks can easily reconstruct the lower frequencies, and the later blocks in the network can use the earlier block outputs to reconstruct higher frequencies. The last block will be the output we use to reconstruct the signal. \\Since we only needed a very shallow and narrow network to output double the inputted frequency, we choose to make the later blocks narrower and shallower than the earlier ones, as their purpose is to act in a similar manner. A graph of the network is presented here:

\begin{figure*}[h]
    \centering
    \includegraphics[width=0.8\textwidth]{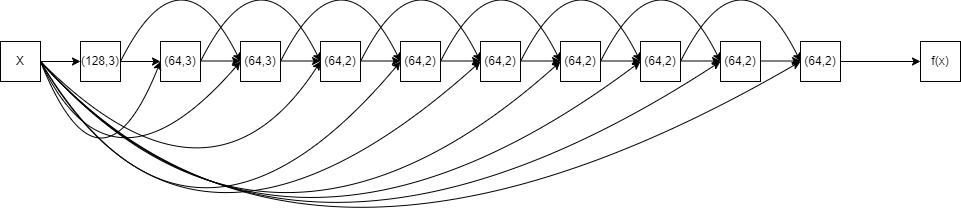}
    \caption{FibonacciNet Architecture. The pair of numbers on each block are its width and depth. the output of each block is used as part of the loss, and the output of the final block gives the output of the network during inference time.}
    \label{fig:fibnet_architecture}
\end{figure*}

Unfortunately, this alone is not sufficient to reconstruct higher frequencies. As the experiment section shows, While the network can easily reconstruct the higher frequencies when the earlier blocks are "spoon-fed" the lower frequencies, this is not the case when attempting to train end-to-end. We therefore suggest a unique training scheme - given the function we're trying to reconstruct, we will want to train by gradually increasing the frequency range of the signal until we are training on the full signal. For this purpose, we will use a low-pass filter with exponentially increasing cutoffs, and use a loss that forces each block to reconstruct the low-pass filtered version of the original function. A low-pass filter takes a signal and removes the fourier frequencies higher than a certain range. By doing this, we will train the network to reconstruct higher and higher frequencies, and the exponential cutoffs will ensure that we quickly will be able to reconstruct the original signal. Formally, if $f$ is the target function, 
\[L=\sum L_i, \quad L_i=MSE(Output_i,Lowpass(f,\mathrm{cutoff}=2^i))\].  

\section{Experiments}

\subsection{Reconstructing a  high-frequency sine wave}
We start by attempting to use the skip-block structure to reconstruct the standard positional encoding scheme:\\ 
$\sin(2^ix), \cos(2^ix)  \forall i \in [10]$

We "spoon-feed" the networks by defining the loss to be \[L=\sum L_i, \qquad L_i=MSE(Output_i,\sin(2^ix))\] where $Output_i$ is the output of the i-th block. The loss forces each block to learn the appropriate frequency - the first block must learn the lowest frequency, and the last one the highest. As expected, since the earlier blocks will be forced into reconstructing the lower frequencies, the later blocks will have an easier time using the output of the earlier blocks to reach the high frequencies. Here we plot the output of the Fibonacci Network:
\begin{figure*}[h]
    \centering
    \includegraphics[height=0.916\textwidth]{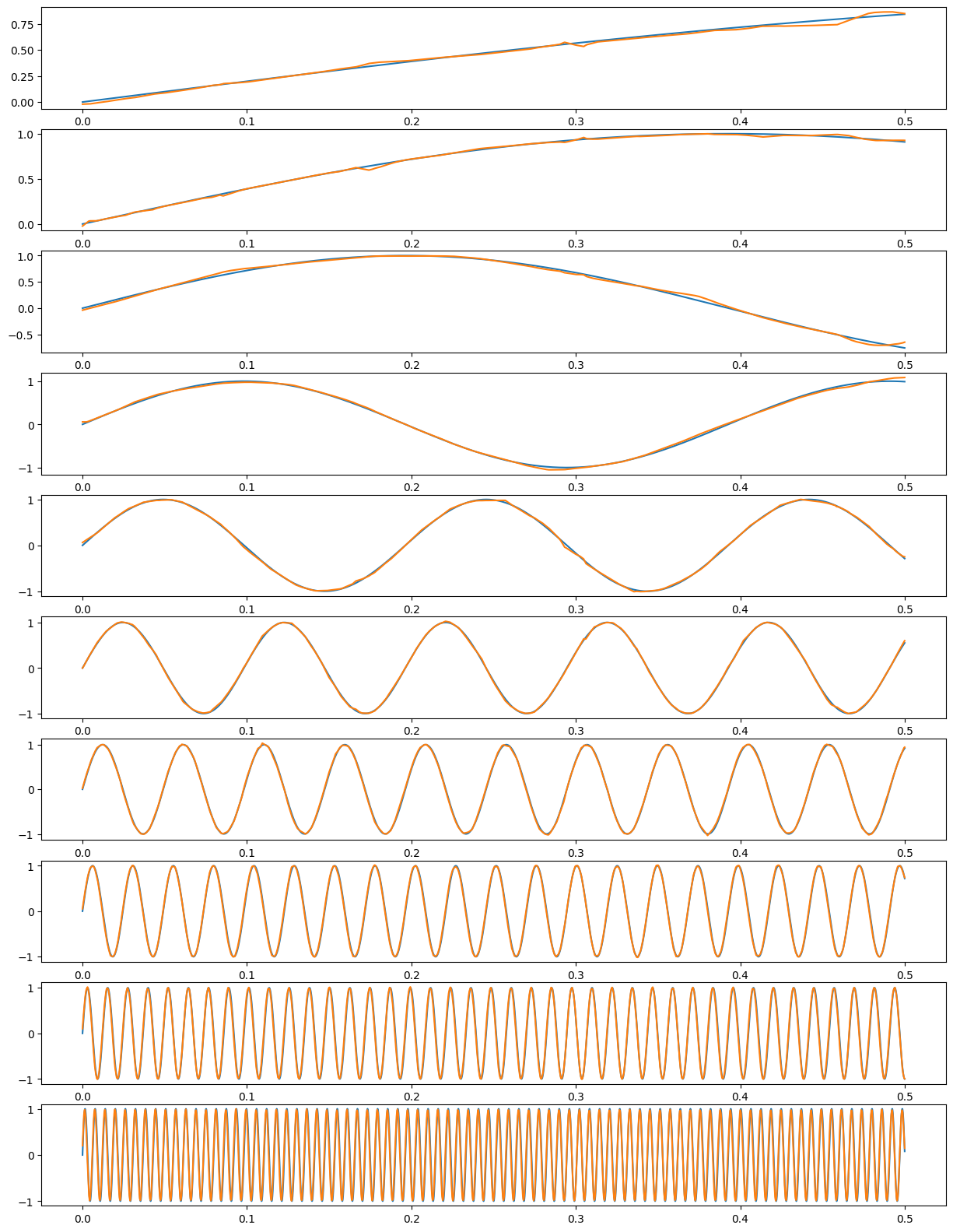}
    \caption{Reconstruction of $\sin(2^ix)$, for $i \in [10]$.}
    \label{fig:pe_reconstruction}
\end{figure*}

This network really shines in the use case of few-shot samplings of very noisy signals. Since the low-pass filter will get rid of much of the noise, the network can reconstruct the signal with relatively good accuracy, even in the case when the noise is incredibly high. In fact, it outperforms positional encoding by a wider and wider margin as noise is added. This is because it is much more robust to noise than positional encoding, which will "hallucinate" high frequencies that aren't in the signal. 

We compare our network architecture with a standard fully-connected network, both with and without positional encoding. 
if a network is able  to not give an unfair advantage to positional encoding, we avoid using sinusoidal functions in our comparisons. Instead, we take a signal that is a cubic spline between random points: we take $x=linspace(0,1,n)$, $y\sim N(0,1)^n$. Then, our function is defined as the interpolation between the points ${(x_i,y_i)}_{i=1}^n$ . We perturb the ground truth signal by adding uniform noise to it, with various levels of noise $\sigma$: $y_i \mapsto y_i+\xi_i, \xi_i \sim U[-\sigma,\sigma]$, and calculate the loss $L_2(N(x),y)$, for the different network architectures N:\newline

\begin{table}
\scalebox{0.94}{
\begin{tabular}{|l|c|c|c|c|c|c|}\hline

\diagbox{NN}{Noise}
          & 0.1 & 0.5 & 1 & 5 & 10 & 20\\\hline
Simple NN  &0.232&0.319&0.458&2.351&6.815&30.887 \\
Simple NN + P.E.&\bf{0.068}  &\bf{0.129}  &\bf{0.266} & 4.172  &16.562&   66.702 \\
Fibnet &0.359&0.391&0.5468&\bf{1.853}&\bf{4.339}&\bf{19.836} \\
\hline
\end{tabular}}
\caption{Comparison of Fibonacci and positional encoding networks on random functions.}
\end{table}

\subsection{Psine function}

Hertz et. al \cite{hertz2021sape}, present an interesting method to increase reconstruction accuracy. They take an encoding that gradually reveals higher frequencies of the encoding as needed and is able  to only use the higher frequencies where necessary in the reconstructed signal. The idea is that this will stop unwanted artifacts from appearing in the reconstructed signal - only the necessary frequencies are used, and none of the ones higher than that. They also take the spatial position into account when choosing the revealed frequencies, since parts of the signal may require the usage of high frequencies that would hurt the performance of the reconstruction in other parts that don't need them (think of a picture in which there is a very complicated pattern in one part of the picture, but the sky is a very simple blue). The example they provide in 1-D is a sinusoid with a frequency that increases exponentially with x, which they call a "Psine" function. We have tested our network on this function with excellent results. However, it appears that both our network and their entire method are not necessary here - a very simple network such as the one they present is sufficient to achieve a very good generalization by simply training for more epochs. As the training is incredibly quick, this is quite easily done. This casts doubt on the efficacy of the gradual revealing of frequencies. Here, we attach a reconstruction of the function using the aforementioned simple network.

\clearpage \bibliographystyle{IEEEbib}
{\footnotesize
\bibliography{refs}}

\begin{thebibliography}{10}

\bibitem{nguyen2015deep}
Anh Nguyen, Jason Yosinski, and Jeff Clune,
\newblock ``Deep neural networks are easily fooled: High confidence predictions
  for unrecognizable images,'' 2015.

\bibitem{park2019deepsdf}
Jeong~Joon Park, Peter Florence, Julian Straub, Richard Newcombe, and Steven
  Lovegrove,
\newblock ``Deepsdf: Learning continuous signed distance functions for shape
  representation,''
\newblock in {\em Proceedings of the IEEE/CVF conference on computer vision and
  pattern recognition}, 2019, pp. 165--174.

\bibitem{szatkowski2022hypersound}
Filip Szatkowski, Karol~J. Piczak, Przemysław Spurek, Jacek Tabor, and Tomasz
  Trzciński,
\newblock ``Hypersound: Generating implicit neural representations of audio
  signals with hypernetworks,'' 2022.

\bibitem{mildenhall2020nerf}
Ben Mildenhall, Pratul~P. Srinivasan, Matthew Tancik, Jonathan~T. Barron, Ravi
  Ramamoorthi, and Ren Ng,
\newblock ``Nerf: Representing scenes as neural radiance fields for view
  synthesis,'' 2020.

\bibitem{saito2019pifu}
Shunsuke Saito, Zeng Huang, Ryota Natsume, Shigeo Morishima, Angjoo Kanazawa,
  and Hao Li,
\newblock ``Pifu: Pixel-aligned implicit function for high-resolution clothed
  human digitization,''
\newblock in {\em Proceedings of the IEEE/CVF international conference on
  computer vision}, 2019, pp. 2304--2314.

\bibitem{niemeyer2020differentiable}
Michael Niemeyer, Lars Mescheder, Michael Oechsle, and Andreas Geiger,
\newblock ``Differentiable volumetric rendering: Learning implicit 3d
  representations without 3d supervision,'' 2020.

\bibitem{chen2019learning}
Zhiqin Chen and Hao Zhang,
\newblock ``Learning implicit fields for generative shape modeling,''
\newblock in {\em Proceedings of the IEEE/CVF Conference on Computer Vision and
  Pattern Recognition}, 2019, pp. 5939--5948.

\bibitem{deng2022nasa}
Boyang Deng, JP~Lewis, Timothy Jeruzalski, Gerard Pons-Moll, Geoffrey Hinton,
  Mohammad Norouzi, and Andrea Tagliasacchi,
\newblock ``Nasa: Neural articulated shape approximation,'' 2022.

\bibitem{henzler2020learning}
Philipp Henzler, Niloy~J. Mitra, and Tobias Ritschel,
\newblock ``Learning a neural 3d texture space from 2d exemplars,'' 2020.

\bibitem{rahaman2019spectral}
Nasim Rahaman, Aristide Baratin, Devansh Arpit, Felix Draxler, Min Lin, Fred~A.
  Hamprecht, Yoshua Bengio, and Aaron Courville,
\newblock ``On the spectral bias of neural networks,'' 2019.

\bibitem{rahimi2007random}
Ali Rahimi and Benjamin Recht,
\newblock ``Random features for large-scale kernel machines,''
\newblock {\em Advances in neural information processing systems}, vol. 20,
  2007.

\bibitem{vaswani2017attention}
Ashish Vaswani, Noam Shazeer, Niki Parmar, Jakob Uszkoreit, Llion Jones,
  Aidan~N Gomez, {\L}ukasz Kaiser, and Illia Polosukhin,
\newblock ``Attention is all you need,''
\newblock {\em Advances in neural information processing systems}, vol. 30,
  2017.

\bibitem{tancik2020fourier}
Matthew Tancik, Pratul Srinivasan, Ben Mildenhall, Sara Fridovich-Keil, Nithin
  Raghavan, Utkarsh Singhal, Ravi Ramamoorthi, Jonathan Barron, and Ren Ng,
\newblock ``Fourier features let networks learn high frequency functions in low
  dimensional domains,''
\newblock {\em Advances in Neural Information Processing Systems}, vol. 33, pp.
  7537--7547, 2020.

\bibitem{jacot2020neural}
Arthur Jacot, Franck Gabriel, and Clément Hongler,
\newblock ``Neural tangent kernel: Convergence and generalization in neural
  networks,'' 2020.

\bibitem{zheng2021rethinking}
Jianqiao Zheng, Sameera Ramasinghe, and Simon Lucey,
\newblock ``Rethinking positional encoding,'' 2021.

\bibitem{hertz2021sape}
Amir Hertz, Or~Perel, Raja Giryes, Olga Sorkine-hornung, and Daniel Cohen-or,
\newblock ``{SAPE}: Spatially-adaptive progressive encoding for neural
  optimization,''
\newblock in {\em Advances in Neural Information Processing Systems},
  A.~Beygelzimer, Y.~Dauphin, P.~Liang, and J.~Wortman Vaughan, Eds., 2021.

\end{thebibliography}

\end{document}